\documentclass{article} 
\usepackage{iclr2025_conference,times}

\usepackage{amsmath,amsfonts,bm}

\def\eqref#1{equation~\ref{#1}}
\def\Eqref#1{Equation~\ref{#1}}

\def\1{\bm{1}}

\DeclareMathAlphabet{\mathsfit}{\encodingdefault}{\sfdefault}{m}{sl}
\SetMathAlphabet{\mathsfit}{bold}{\encodingdefault}{\sfdefault}{bx}{n}

\newcommand{\Var}{\mathrm{Var}}

\usepackage{hyperref}
\usepackage{url}

\title{Hybrid Generative Modeling for Incomplete Physics: Deep Grey-Box Meets Optimal Transport}

\iclrfinalcopy
\author{Gurjeet Sangra Singh$^{1,2}$,~~ Maciej Falkiewicz$^{1,2}$,~~Alexandros Kalousis$^{2}$\\
Department of Computer Science \\
$^{1}$University of Geneva, $^{2}$HES-SO/HEG Genève, Switzerland \\
\texttt{\{gurjeet.singh, maciej.falkiewicz\}@etu.unige.ch} \\
\texttt{\{gurjeet.singh, maciej.falkiewicz, alexandros.kalousis\}@hesge.ch} 
}

\newcommand{\vect}[1]{\boldsymbol{#1}}
\DeclareMathOperator*{\arginf}{arginf}
\DeclareMathOperator*{\argsup}{argsup}
\definecolor{darkgreen}{HTML}{2E7D32}
\definecolor{salmon}{rgb}{0.98,0.50,0.45}
\definecolor{grey}{gray}{0.6}

\usepackage{enumitem}
\usepackage{amstext}
\usepackage{amsmath}
\usepackage{xcolor}
\usepackage{comment}
\usepackage{tikz}
\usetikzlibrary{decorations.pathmorphing}
\usepackage{booktabs} 
\usepackage{caption} 
\usepackage{graphicx} 
\usepackage[linesnumbered,ruled,vlined]{algorithm2e}
\usepackage{graphicx}
\usepackage{subcaption}
\usepackage{cleveref}
\usepackage{amsfonts}

\begin{document}
 
\maketitle

\begin{abstract}
Physics phenomena are often described by ordinary and/or partial differential equations (ODEs/PDEs), and solved analytically or numerically. Unfortunately, many real-world systems are described only approximately with missing or unknown terms in the equations. This makes the distribution of the physics model differ from the true data-generating process (DGP). Using limited and unpaired data between DGP observations and the imperfect model simulations, we investigate this particular setting by completing the known-physics model, combining theory-driven models and data-driven to describe the shifted distribution involved in the DGP. We present a novel hybrid generative model approach combining deep grey-box modelling with Optimal Transport (OT) methods to enhance incomplete physics models. Our method implements OT maps in data space while maintaining minimal source distribution distortion, demonstrating superior performance in resolving the unpaired problem and ensuring correct usage of physics parameters. Unlike black-box alternatives, our approach leverages physics-based inductive biases to accurately learn system dynamics while preserving interpretability through its domain knowledge foundation. Experimental results validate our method's effectiveness in both generation tasks and model transparency, offering detailed insights into learned physics dynamics.
\end{abstract}
\begin{figure}[h]
    \centering
    \includegraphics[width=0.6\linewidth]{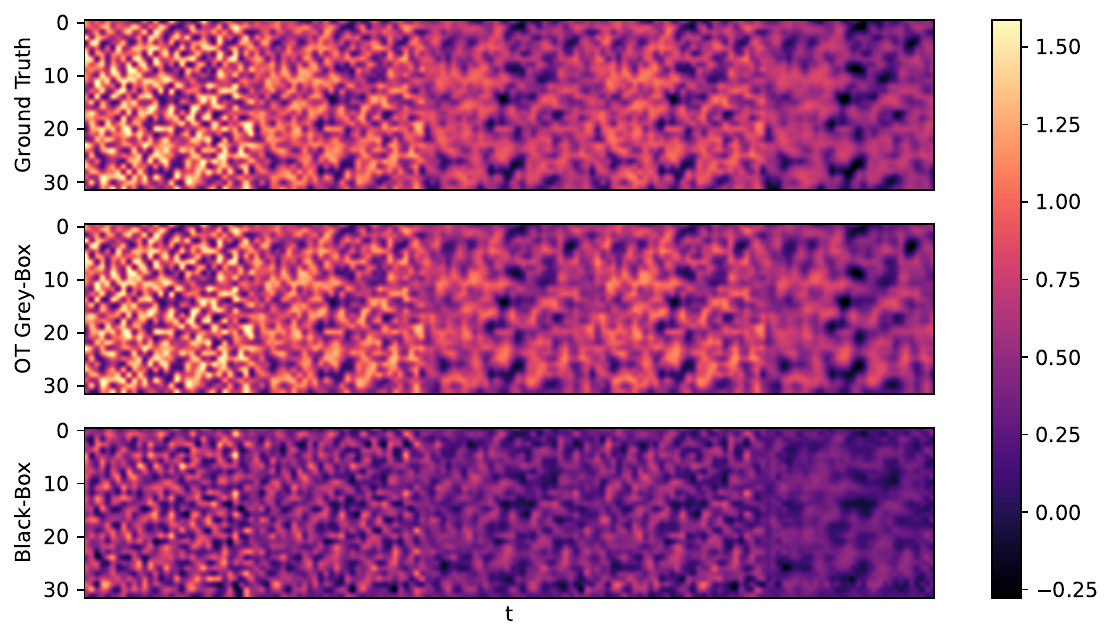}
    \caption{\textbf{Example of a Reaction Diffusion generation}. \small A visual comparison of the OT black-box and grey-box model generation. The grey-box model shows a generation that closely matches the ground truth.}
\end{figure}

\section{Problem Setting}
Consider an unknown data-generating process (DGP) underlying the real world observations, and assume that we have access to some physics-based model $f_p: \mathcal{X}\times \Theta \mapsto \mathbb{R}^d$, which provides only an imperfect description of the DGP, where $\mathcal{X} \subseteq \mathbb{R}^d, \Theta \subseteq \mathbb{R}^p$. We focus on ODE/PDE-based models as our physics descriptions $f_p$.
Let the DGP samples be independent identically distributed (iid), we denote the collections of these samples by $\mathcal{D}^y=\{\vect{y}_i\}_{i=1}^M$. Each $\vect{y}_i$ sample corresponds to a time series $\vect y =[\vect y(0), ..., \vect y(T)] $ with $\vect{y}(t) \in \mathcal{Y} \subseteq \mathbb{R}^d$ for each time step $t \in \{0, \dots T\}$. These trajectories are sampled from an unknown \textit{target} distribution $\vect{y} \sim \nu(\vect{y}) \in \mathcal{P}(\mathcal{Y}^{T+1})$, and $\mathcal{P}(\mathcal{Y}^{T+1})$ is a set of all probability measures over the $\mathcal{Y}^{T+1}$ trajectory space. 
Simulating the physics-based model $f_p$ is possible by a simulator $S_{f_{p}}: (\vect x_0, \vect{\theta}) \mapsto \vect{x}$, given its initial conditions $\vect{x}_0 \in  \mathbb{R}^d$ and physics parameters $\vect{\theta} \in \Theta \subseteq \mathbb{R}^p$ it produces a simulated observation $\vect{x} \in \mathcal{X}^{T+1}$, where $\vect{x}=[\vect x(0), \dots, \vect x(T)]$ with $\vect x(0)=\vect x_0$ its initial condition, and $\vect x(t) \in \mathbb{R}^d$. We can generate simulated observations by sampling from the priors $p(\vect \theta)$ and $p(\vect x_0)$, and running the samples through the simulator. 
We denote the intractable probability distribution of the simulator by $\vect{x} \sim p(\vect{x} | \vect{x}_0, \vect{\theta})$, and the
pair of parameters and model simulations by $\mathcal{D}^x = \left\{(\vect{\theta}_i, \vect{x}_i)\right\}_{i=1}^{M}$. We refer to the marginal distribution of the simulations as the \textit{source} distribution $\mu(\vect x) \in \mathcal{P}(\mathcal{X})$, where $\mu(\vect x) = \int p(\vect{x} | \vect{x}_0, \vect{\theta}) p(\vect{\theta}) p(\vect{x}_0) d\vect{\theta} d\vect{x}_0$. 
We do not have access the pair $(\vect x, \vect y)$ association between source and target distribution samples.

\textbf{Imperfect model and misspecification}.
We assume that the physics model $f_p$ is a simplified description of the real-world phenomena and only partially describes the observations. Therefore the \textit{marginal of the simulated model} does not coincide with the distribution of the DGP:
\begin{align}
    \int p(\vect{x} | \vect{x}_0, \vect{\theta}) p(\vect{\theta}) p(\vect{x}_0) d\vect{\theta} d\vect{x}_0 = \mu(\vect{x}) &\neq \nu(\vect{y}),
\end{align}
We say thus that the physics model is imperfect or misspecified with respect to the true DGP.

\textbf{Objective}. One of the main goals of our work is to solve unpaired translation (one-to-one and one-to-many translations) between simulations $\vect{x}_i$ of the $f_p$ imperfect model and real-world observations $\vect{y}_i$, by preserving the correct usage of the input and parameters of the physics model. In this setting, it translates to learning a conditional distribution $p_\phi (\vect{y} | \vect{x}, \vect\theta,  \vect z)$ such that its marginal approximates the target distribution $\nu(\vect y)$:
\begin{align}
\label{eq:learned-prob-model}
    \int p_\phi (\vect{y} | \vect{x}, \vect\theta \vect, \vect z) p(\vect{x} | \vect{x}_0, \vect{\theta}) p(\vect{\theta}) p(\vect{x}_0) p(\vect{z}) d\vect{\theta} d\vect{x}_0 d\vect{z} = \hat\nu(\vect{y}) \approx \nu(\vect{y}),
\end{align}
where $\vect z \in \mathbb{R}^d$ is a latent variable, extending the model to solve one-to-many translation maps. 

\begin{figure}[t!]
    \centering
    \resizebox{0.6\textwidth}{!}{
    \begin{tikzpicture}
        \node[circle, draw, fill=grey!30, minimum size=1cm] (init) at (3.2, 4) {$\vect{x}_0$};
        \node[circle, draw=blue, dash pattern=on 1pt off 1.5pt, line width=1pt, minimum size=1cm] (theta) at (2,2) {$\vect \theta$};
        \node[circle, draw=blue, dash pattern=on 1pt off 1.5pt, line width=1pt, minimum size=1cm] (z) at  (5,4) {$\vect z$};
        \node[circle, draw, fill=grey!30, minimum size=1cm] (x) at (5,2) {${\vect{y}}$};
        \node[circle, draw, fill=grey!30, minimum size=1cm] (x_latent) at (1,4) {${\vect x}$};
        
        \draw[->] (z) -- (x);
        \draw[->] (theta) -- (x);
        
        \draw[->] (init) -- (x);
        \draw[->] (theta) -- (x_latent);
        \draw[->] (init) -- (x_latent);
        \draw[->] (x_latent) -- (x);
        
        \draw[black, thick, dashed] (0.3, 1.3) rectangle (4.0, 5.0);
        \draw[black, thick] (4.3, 1.3) rectangle (5.7, 5.0);
        
        \node[anchor=north west] at (0.3, 5.) {\small Physics model};

        \node[anchor=north west] at (7.5, 5.0) {\small \textbf{Legend:}};
        \node[circle, draw, fill=grey!30, minimum size=0.3cm] at (7.7, 4.2) {};
        \node[anchor=west] at (8.0, 4.2) {\small Observed};
        \node[circle, draw=blue, dash pattern=on 1pt off 1.5pt, line width=0.8pt, minimum size=0.3cm] at (7.7, 3.8) {};
        \node[anchor=west] at (8.0, 3.8) {\small Latent};
    \end{tikzpicture}
    } 
    \caption{Probabilistic graphical model of the hybrid model}
    \label{fig:PGM}
\end{figure}
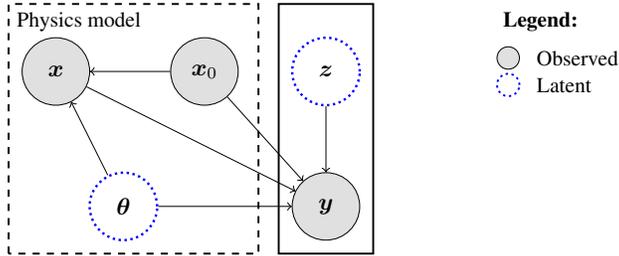

In our work we characterize $p_\phi (\vect{y} | \vect{x}, \vect\theta)$ via a conditional map $T_\phi(\vect{\theta}, \vect z): \vect{x} \rightarrow \vect{y}$, which maps source samples $\vect x \sim \mu(\vect x)$ to target samples $\vect y \sim \nu(\vect y)$ without pair associations. We remark that learning the map $T_\phi(\vect x ; \vect{\theta}, \vect z)$ conditioned on $\vect \theta$ is crucial in our setting since it allows the physics parameter inference, which is a common downstream task. A central tool in addressing the setting we just described will be optimal transport, motivated by its ability to preserve the structure of input sample with minimum change effect, a desired property in unpaired learning and tightly linked to the principle least action of the physics dynamic. A schematic formulation of the probabilistic model is shown in Figure \ref{fig:PGM}.

\section{Methodology}
For solving the unpaired translation problem, our method learns the optimal conditional mapping $T_\phi(\vect x;\vect \theta, \vect z)$, constrained by the $f_p$ physics model. A careless choice of the learned model, especially when considering a fully data-driven approach, can lead to inefficient and potentially wrong solutions because of several significant key challenges in the problem. First, while the $T_\phi(\vect x;\vect \theta, \vect z)$ can approximate the marginal distributions if no explicit physics constraints are used, they can fail to preserve physical interpretability, as studied in the work of \cite{naoya_2021}. This is problematic because aligning the marginal distributions alone in \Eqref{eq:learned-prob-model} does not guarantee the correct usage of conditional parameters, which can result in incorrect physics predictions with respect to its parameters. 

Second, when considering deep grey-box models of the form $f_\phi \circ f_p$ (see Appendix \ref{app:background-gb} for details), the high expressiveness of deep neural networks $f_\phi$ may cause the learned model to bypass or override the physics-based component $f_p$. The latter has been studied by \cite{naoya_2023} under supervised regression problems, concluding that regularization of the learned network is indeed needed to ensure physics constraints in hybrid architectures and prevent $f_p$ from being overwritten.

These challenges highlight the need for a principled approach that can both align marginal distributions while respecting physical constraints. This motivates our use of Optimal Transport (OT) methods as a central tool of our solution.

We employ Optimal Transport (OT) methods to align the marginal distributions \footnote{Readers are encouraged to consult the appendix on the Optimal Transport (OT) method in the background section \ref{app:background-ot} to ensure familiarity with the notation used}. The OT framework naturally incorporates the principle of least action in the learned model when transforming the source distribution to the target distribution. 
In our solution, we learn a \textit{conditional optimal stochastic map function} by solving a weak-OT problem (\Eqref{eq:weak-ot-2} and its dual form \Eqref{eq:weak-ot-dual}) using the Maximin formulation proposed by \cite{korotin2023kernel}, which provides a solution to the OT in the data space. Within our conditional learning setting, the Maximin problem reformulates to:
\begin{align}
\label{eq:kernel-maximin}
    W_{k, \gamma}(\mu, \nu) = \sup_{f} \inf_{T} \int_{\mathcal{X}} \left(C_{k, \gamma}\left(\vect x, T_{\vect x}\#\eta\right) - \int_{\mathcal{Z}} f(T_{\vect{x}}(\vect \theta, \vect z) )d\eta(\vect z) \right) d\mu(\vect x) + \int_{\mathcal{Y}} f(\vect y) d\eta(\vect y)
\end{align}
where $\eta \in \mathcal{P}(\mathcal{Z}), \mathcal{Z} \subseteq \mathbb{R}^d$ is a latent distribution from which sampling is easy (e.g gaussian, uniform distributions). $T(\vect \theta, \vect z): \vect x  \mapsto \vect y$ maps from source to  the target samples and $T_{\vect x}\#\eta$ denotes its pushforward measure of $\eta$ under the map $T_{\vect x}$. $f: \mathbb{R}^d \mapsto \mathbb{R}$ is a potential function satisfying the properties of \Eqref{eq:weak-ot-dual}. Finally, $C_{k, \gamma}(\vect x, \nu)$ extends the weak-OT cost of \Eqref{eq:weak-ot-2} with a kernel weak quadratic cost:
\begin{align}
\label{eq:kernel-cost}
    C_{k, \gamma}(\vect x, \nu) &=  \frac{1}{2} k(\vect x, \vect x) +  \frac{1-\gamma}{2} \int_{\mathcal{Y}} k(\vect y, \vect y)d\nu(\vect y) + \frac{\gamma}{2} \int_{\mathcal{Y} \times \mathcal{Y}} k(\vect y, \vect y') d\nu(\vect y) d\nu(\vect y'),
\end{align}
where $C_{k, \gamma}$ is defined in the Hilbert Space $\mathcal{H}$, and $k:\mathcal{Y} \times \mathcal{Y} \mapsto \mathbb{R} $ is a continuous characteristic Positive Definite Symmetric (PDS) kernel \footnote{A PDS kernel $ k : \mathcal{Y} \times \mathcal{Y} \mapsto \mathbb{R}$ is characteristic if the kernel mean embedding $\mu \in \mathcal{P}(\mathcal{Y}).\; \mu \mapsto u(\mu) := \int_\mathcal{Y} u(\vect y)d\mu(\vect y) \in \mathcal{H} $} for mapping features. Solving \Eqref{eq:kernel-maximin} with the above related cost function, and $\gamma \in (0,1]$, guarantees the optimality of stochastic functions in all optimal saddle points for any optimal potential $f^* \in \argsup_f \inf_{T} W_{k, \gamma}$ (Theorem 2 of \cite{korotin2023kernel}). That is, $T^* \inf \arginf_T  W_{k, \gamma}(\mu, \nu; f^*)$ yields a stochastic OT map $T^*$. 

Practically, the solution of the dual optimization is achieved using an adversarial method, by parametrizing $T=T_\phi$ and $f = f_\psi$ with a neural network. We alternate the optimization of the parameters of the conditional generator/map $T_\phi(\vect x,\vect \theta, \vect z)$ and its potential function $f_\psi$, where $\vect y$ is sampled from the DGP, and the  $\vect{x}$ source sample is achieved by simulating the imperfect model $S_{f_p}: (\vect x_0, \vect \theta) \mapsto \vect x$. We enforce on the potential function $f_\psi$ some typical penalty loss of \cite{wgans-2017b, rout2022generative} to satisfy the smoothness of not very rapid growth required by the OT formulation of \Eqref{eq:weak-ot-dual}. When solving deterministic map one-to-one translation, we remove the latent $\vect z$ variable and set $\gamma=0$. Instead, when solving generative one-to-many maps, we define a latent variable  $\vect z$ from a discrete or continuous space and set $\gamma=1$, inducing the algorithm to associate multiple maps for each input.
In our experiments, we chose the distance-based kernel proposed by \cite{korotin2023kernel} in the weak OT cost. We report the actual implementation of the algorithm in Section \ref{app:alg} of the Appendix, and make the code available on our GitHub repository \footnote{Code repository: \href{https://github.com/DMML-Geneva/ot-deep-grey-box-model}{github.com/ot-deep-grey-box-model}}.

While the OT framework described above effectively addresses the challenge of aligning marginal distributions, it is only part of our complete solution strategy.

\textbf{Beyond Marginal Distribution Alignment}. Apart from aligning marginal distribution from source to target, another fundamental challenge in our task is to validate the correct usage of conditional physics model parameters. Learning a conditioned model on $\vect \theta$ is crucial for downstream tasks, such as inverse problems in physics, where the parameter inference determines the expressibility and validity of the model in representing real observations. For this reason, we assess the capabilities of the learned models to correctly use the conditional parameters $\vect \theta$ and the generated trajectory $\vect y$, by testing them on the underlying true joint distribution $\pi(\vect x, \vect\theta, \vect y)$, we provide a more detailed description in the evaluation method described in Section \ref{sect:eval}.

\textbf{Physics-Guided Optimal Transport Integration}. We bridge the OT formulation of \Eqref{eq:kernel-maximin} with hybrid grey-box models to address the correct usage of physics parameters in the optimal transport solution. We incorporate deep grey-box models as conditional maps in the OT algorithm, where $T_\phi=\mathcal{T}(f_p, T_\phi; \vect x,\vect \theta)$ follows \Eqref{eq:phys-neural-ode}. 
This integration serves two purposes: first, it leverages the known structure of $f_p$ by incorporating the physics dynamics of the known model, minimizing the intervention of the learned component $T_\phi$ in our generative model. Second, it secures the usage of the conditional parameters $\vect \theta$ by $f_p$, which acts as a soft constraint on the model by narrowing the solution space of the pushforward map during optimization to find the optimal stochastic map solution efficiently.

To demonstrate the efficiency of our hybrid model solution, we validate our approach through extensive evaluation methods on numerical experiments, showing that our method achieves superior performance and generates reliable solutions in learning the correct usage of the input and physics parameter. Our comparative analysis spans various physics tasks, examining different levels of physics constraint encoding, ranging from our proposed method to fully data-driven black-box models.

Additionally, for hybrid models, we conduct a component analysis of the grey-box model by examining the accuracy of the learned physics equation at the functional level. This demonstrates the capability of grey-box models to offer deeper insight into the prediction process, ultimately leading to more trustworthy solutions.

\section{Related work}
Deep grey-box models are steadily gaining their attention in scientific deep learning applications, particularly those involving physical dynamical systems such as PDEs \cite{yin_2021,kon2022,Mehta2020NeuralDS,agarwal2024hybrid}. This growing interest has been driven by recent advances in differentiable solvers that enable scalable backpropagation \cite{Chen_2021}. 

Recent studies have demonstrated the potential of these models in several key areas: as efficient generative models for learning dynamical systems \cite{verma2024climode, thoreau2023p3vae, naoya_2021, wehenkel2023robust}, and providing more reliable parameter inference in physical systems \cite{naoya_2021,wehenkel2023robust}, allowing also inspection and interpretability of each component of the model. Although hybrid models have been studied for many years \cite{greybox1992,greybox1994a,greybox1994b}, the study of regularized learning methods involving deep learning to control trainable components remains limited \cite{naoya_2023}.

Our work extends the studies of \cite{naoya_2021} and \cite{wehenkel2023robust} by introducing a novel application of hybrid generative models in an unsupervised setting. Specifically, we learn to complete dynamics from a source distribution based on incomplete physics models to a target distribution derived from the data-generating process (DGP), using conditional generative models by combining Optimal Transport (OT) method and hybrid models. While related, our approach differs from previous works \cite{naoya_2021, wehenkel2023robust} where the probabilistic model comes as a VAE architecure, which learns to complete the physics model by a decoder grey-box model and an encoder network that infers the physics $\vect \theta \sim p(\vect \theta | \vect y)$ parameters. These methods requires multiple regularizations of the encoder and decoder models to ensure the correct usage of the $f_p$ model. Instead, our method focuses specifically in the generative setting - without involving any physics parameter inference. Our method comes with a neat formulation and  well-founded by the OT theory, where the OT ground-cost enforce the minimal changes from the source distribution $\mu(\vect x)$ using the incomplete $f_p$ physics model, without the need of any regularisations and augmentation similar to \cite{naoya_2021, wehenkel2023robust}. Since we only focuses on the generative task where we aim to correct and complete the physics model, we establish our baseline by comparing Wasserstein-GAN \cite{wgans-2017, wgans-2017b} and OT methods, evaluating the effectiveness of the methods and between grey-box against black-box models.

Additionally, we conduct in our work some analysis on the component of  grey-box model to analyse how OT methods utilize incomplete physics models and the learned components to complement rather than override the physics model, similar to the investigation of \cite{naoya_2023}.
Other methods incorporating physics knowledge into deep learning models, such as Physics-Informed Neural Networks (PINNs) \cite{raissi2019physics}, are of great interest but remain outside the scope of this study due to their different methodological approaches and objective.

\section{Evaluation - A Comprehensive Evaluation Framework for Physics Models}
\label{sect:eval}
Evaluating the models only on the generative performance between $\nu(\vect y)$ and the approximated $\hat{\nu}(\vect y)$ distribution does not to assess the crucial relationship between the input of the physics model and their corresponding outputs, leading to inconsistent or physically invalid results in downstream applications. Therefore, in our setting, a correct evaluation strategy should first be conducted on the numerical experiments where the underlying true joint distribution $\pi(\vect x, \vect\theta, \vect y)$ between input  simulations, physics parameter, and target samples is available through samples.

Furthermore, we are also interested in comparing various modelling approaches for incorporating physics into machine learning models—ranging from grey-box methods that explicitly utilize the $f_p$ physical model to purely data-driven black-box models. For this reason, we dedicate in our experiments a study of the  comparison among different model 

\textbf{Evaluation setting}. We test our models on two main experiment settings to assess the strengths and weaknesses of the different approaches, by assessing their capabilities in learning from simple deterministic maps to complex stochastic solutions. We consider first a one-to-one translation problem where the solutions lie on a deterministic mapping solution, $\forall \vect (\vect \theta, \vect x) \text{ s.t. } T(\vect x, \vect \theta) \mapsto \vect y$ is unique. Instead for the stochastic setting, the solutions from source to target samples do not have a one-to-one unique map but present a one-to-many mapping solution. 

In our synthetic experiments, the DGP is given by a completed phyiscs model $f_c = f_q \circ f_p$, where $f_q$ is the true missing term and potentially incorporates some latent physics variables $\vect \theta'$. Thus, $\vect y$ samples of the DGP are generated by simulating the complete $f_c$ through the $S_{f_c}$ simulator. Finally, the evaluated joint samples $(\vect x, \vect\theta, \vect y) \sim \pi(\vect x, \vect\theta, \vect y)$ are generated by first simulating on the complete model $S_{f_{c}}$ and later on by simulating the incomplete model $S_{f_{p}}$ from the same $\vect x_0$ and $\vect \theta$ prior samples. When considering the one-to-many stochastic setting, the complete model $S_{f_{c}}$ presents some stochastic latent physics variables $\vect \theta'$, thus multiple $\vect y_j$ are generated from the same pair $\vect x_0$ and $\vect \theta$ allowing us to assess the generalization of the model in learning the multiple mapping solutions. We denote the test data by the following set $\mathcal{D_{\text{test}}}=\left\{ \left(\vect x_i, \vect \theta_i, \left\{\vect y_j \right\}_{j=1}^M \right) \right\}_{i=1}^N$.


\textbf{Performance Metrics}
To compare the performance of the generative model, we evaluated all learned models using the exact input and parameter of $\mathcal{D}_{\text{test}}$ pairs and computed a discrepancy measure among samples.
The metrics involved depend on two main configurations:
\begin{enumerate}[label={(\roman*)}]
    \item Deterministic Setting: For one-to-one mapping problems, we calculated the Absolute Square Error (ABS) or Normalized Root Mean Square Error (N-RMSE) between predictions and targets generated from the same input pairs $(\vect \theta, \vect x)$.
    \item Stochastic Setting: For one-to-many mapping problems, we employed:
    \begin{itemize}
        \item Classifier Two-Sample Test (C2ST) \cite{c2st-lopez-paz2017} by evaluating on joint samples from $\pi(\vect \theta, \vect x, \vect y)$
        \item Maximum Mean Discrepancy (MMD) with aggregated kernels \cite{gretton12a, schrab2021mmd} evaluated on joint samples $\pi(\vect \theta, \vect x_0,\vect y)$\footnote{To calculate the MMD, we consider $(\vect x_0, \vect \theta)$ as these values completely determine $\vect x$ by $f_p$ according to our ODE/PDEs, avoiding the complexities of a high-dimensional space.}
    \end{itemize}
\end{enumerate}

For the MMD, we take the kernel to be the exponentiated-quadratic kernel $k_\sigma(\vect s, \vect s') = \exp (-||\vect s - \vect s'||^2 \sigma^{-2})$, and set its length-scale $\sigma$ using the median heuristic of norm-2 distances for all pairs of the set data points $\{\vect{s}_i\}_{i=1}^M$ \cite{gretton12a}, and afterwards we constructed the aggregated kernels using the adaptive rule suggested by \cite{schrab2021mmd}. Instead, for the C2ST we computed the Accuracy score given by the classifier, and for each task, we designed its neural network architecture. 
These metrics provide a comprehensive evaluation of the model's ability to capture complex relationships in the data, ensuring robust validation for downstream applications.


\section{Experiments}
\begin{table}[t]
\caption{\small \textbf{A comparison of grey-box and black-box methods using the joint evaluation.} The reference scores are computed using bootstrapping on different test set splits.}
\label{tab:grey-vs-black-box}
\centering
\resizebox{\textwidth}{!}{%
\begin{tabular}{lcccccccc}
    \toprule
        \multicolumn{9}{c}{\textbf{(a) One-to-One Mapping Tasks}} \\ \midrule
        Method & \multicolumn{2}{c}{\textbf{\textcolor{darkgreen}{OT}-\textcolor{blue}{GB}}} & \multicolumn{2}{c}{\textbf{WGAN-\textcolor{blue}{GB}}} & \multicolumn{2}{c}{\textbf{\textcolor{darkgreen}{OT}-BB}} & \multicolumn{2}{c}{\textbf{WGAN-BB}} \\
        \cmidrule(r){2-3} \cmidrule(lr){4-5} \cmidrule(lr){6-7} \cmidrule(l){8-9}
        Experiment & \textbf{N-RMSE} & \textbf{ABS. ERR} & \textbf{N-RMSE} & \textbf{ABS. ERR} & \textbf{N-RMSE} & \textbf{ABS. ERR} & \textbf{N-RMSE} & \textbf{ABS. ERR} \\ \midrule
        Pendulum-1-m  & $\vect{0.0096} \pm \vect{0.0045}$ & $\vect{0.0066} \pm \vect{0.0021}$ & $0.0143 \pm 0.0037$ & $0.0083 \pm 0.0017$ & $0.0887 \pm 0.0026$ & $0.050 \pm 0.003$ & $0.41 \pm 0.15$ & $0.2897 \pm 0.0796$ \\
        Adv-diff-1-m & $\vect{0.0217} \pm \vect{0.003}$ & $\vect{0.0171} \pm \vect{0.0026}$ & $\vect{0.0200} \pm \vect{0.007}$ & $\vect{0.0165} \pm \vect{0.0064}$ & $0.0528 \pm 0.0028$ & $0.0387 \pm 0.0020$ & $0.0340 \pm 0.0028$ & $0.0277 \pm 0.0035$   \\
        React-diff-1-m & $\vect{0.0355} \pm \vect{0.0029}$ & $\vect{0.0328} \pm \vect{0.0008}$ & $\vect{0.0312} \pm \vect{0.0004}$ & $\vect{0.0330} \pm \vect{0.0022}$ & $0.0705 \pm 0.0250$ & $0.0724 \pm 0.0269$ & $0.1109 \pm 0.0106$& $0.1083 \pm 0.0120$ \\ 
    \bottomrule
\end{tabular}
}
\vspace{0.5em} 

\resizebox{\textwidth}{!}{%
\begin{tabular}{lcccccccccc}
    \toprule
        \multicolumn{11}{c}{\textbf{(b) One-to-Many Mapping Tasks}} \\ \midrule
        Method & \multicolumn{2}{c}{\textbf{\textcolor{darkgreen}{OT}-\textcolor{blue}{GB}}} & \multicolumn{2}{c}{\textbf{WGAN-\textcolor{blue}{GB}}} & \multicolumn{2}{c}{\textbf{\textcolor{darkgreen}{OT}-BB}} & \multicolumn{2}{c}{\textbf{WGAN-BB}} & \multicolumn{2}{c}{\textbf{Reference}} \\
        \cmidrule(r){2-3} \cmidrule(lr){4-5} \cmidrule(lr){6-7} \cmidrule(lr){8-9} \cmidrule(l){10-11}
        Experiment & \textbf{MMD} & \textbf{C2ST} & \textbf{MMD} & \textbf{C2ST} & \textbf{MMD} & \textbf{C2ST} & \textbf{MMD} & \textbf{C2ST} & \textbf{MMD} & \textbf{C2ST} \\ \midrule
        Pendulum-2-m  & $\vect{0.010} \pm \vect{0.003}$ & $\vect{0.72} \pm \vect{0.06}$ & $0.024 \pm 0.0001$ & $0.89 \pm 0.06$ & $0.078 \pm 0.017$ & $0.94 \pm 0.08$ & $0.124 \pm 0.027$ & $0.98 \pm 0.01$ & $\vect{0.016} \pm \vect{0.0036}$ & $\vect{0.51} \pm \vect{0.03}$ \\
        Pendulum-m-m  & $\vect{0.011} \pm \vect{0.002}$ & $\vect{0.77} \pm \vect{0.02}$ & $0.022 \pm 0.0003$ & $0.87 \pm 0.06$ & $0.029 \pm 0.001$ & $0.99 \pm 0.001$ & $1.064 \pm 1.388$ & $0.99 \pm 0.003$ & $\vect{0.011} \pm \vect{0.004}$ & $\vect{0.51} \pm \vect{0.02}$ \\
        Adv-diff-m-m  & $\vect{0.0396} \pm \vect{0.0032}$ & $\vect{0.57} \pm \vect{0.03}$ & $0.0430 \pm 0.00463$ & $0.73 \pm 0.04$ & $0.0512 \pm 0.005$ & $0.74 \pm 0.13$ & $0.131 \pm 0.017$ & $0.99 \pm 0.01$ & $\vect{0.0081} \pm \vect{0.0038}$ & $\vect{0.50} \pm \vect{0.02}$ \\
    \bottomrule
\end{tabular}
}
\end{table}
\label{sect:exps}

We analyse the baseline methods and our solution on different numerical experiments, following benchmarks used in previous works \cite{naoya_2023, wehenkel2023robust, naoya_2021}, and adapt to our setting. In this section, we first present the configuration setting for each task, and then the results of our models and the baseline models. All task experiments presented below are evaluated on a test dataset whose size is 20\% or 10\% of the training observation $|Y|$ set size. Detailed technical information for each task and trained model is given in Appendix \ref{app:exps}.

\subsection{Unpaired one-to-one physics translation mapping tasks}
\textbf{Damped Pendulum}. We consider as a DGP a non-linear damped pendulum:
\begin{equation}
\label{eq:pendlum-fric}
    \underbrace{\frac{\mathrm{d}^2 }{\mathrm{d} t^2} x(t)+\omega^2 \sin x(t)}_{f_p} + \xi\frac{x(t)}{dt}=0,
\end{equation}
with a fixed damping coefficient $\xi=1.2$, and initial pendulum's angle and  angular velocity sampled from a uniform distribution, respectively  $\omega \sim \mathcal{U}(0.785, 3.14)$, $x(0) \sim \mathcal{U}(-1.57, 1.57)$. For training our model, we used $|Y|=1000$ sample trajectories of length $\tau=50$ (time interval $[0,50]$ and$\Delta t=0.1$) as a target distribution (damped pendulum), generated using the Runge-Kutta method of order 4. The source  distribution samples are generated by the friction-less pendulum model $f_p$, with the same ODE solver configuration. We used $|X| = 10^6$ samples of the source distribution as a constraints simulation budget.

\textbf{Reaction diffusion} In this task we study a high dimensional problem ($\vect x, \vect y \in \mathbb{R}^{2\times32\times32\times 15}$), which involves a two-dimensional reaction-diffusion PDE of FitzHugh-Nagumo mode:
\begin{equation}
\frac{du}{dt} = a\Delta u + u - u^3 -v - k, \quad \frac{dv}{dt} = b\Delta v + u -v .
\end{equation}
$\Delta$ describes the Laplace operator, $a, b$ are the parameter of the system, and $u(0), v(0)$ the initial conditions. To generate the source samples, the model is misspecified by the $k$ parameter, which is not included in the imperfect model. The simulations result in a $[u(t),v(t)] \in \mathbb{R}^{2 \times 32\times32}$ states' space, evaluated on $15$ episodes of the dynamic system. We used $|Y|=500$ target samples and $|X|=1.5 \cdot 10^6$  source budget samples. A similar setting has also been considered by \cite{wehenkel2023robust, naoya_2023}, a detailed description is presented in Appendix \ref{app:exps}.

\textbf{Advection Diffusion}. To produce the dataset, we used the following PDE: 
\begin{equation}
\label{eq:adv-diff}
    \underbrace{\frac{d T(s,t)}{dt} - \alpha \frac{d^2 T(s,t)}{ds^2}}_{f_p} + \beta \frac{dT(x,s)}{ds} = 0,
\end{equation}
with $s$ being a 1-D spatial dimension diffusion on 20-point even grid. We evolved the system for $50$ time steps ($t=50$), with a time-step size of $\Delta t=0.02$. The simulation corresponds to a temperature matrix of $\vect T \in \mathbb{R}^{20\times 50}$. The boundary conditions and initial conditions are defined as: $T (0, s) = c \sin(\pi s/2), \;T (t, 0) = T (t, 2) = 0$, where $c$ is sampled randomly from the range $[0.5, 1.5]$.
The known $f_p$ physics model of the source simulations does not include the advection term $\beta \frac{dT(x,s)}{ds}$. Instead, the $\alpha$ diffusion coefficient is sampled from a $\mathcal{U}(10^{-2}, 10^{-1})$.  For the one-to-one translation map, the DGP samples include the $\beta$ term, fixed to $\beta=0.1$, with $|Y|=1000$ target samples and $|X|=10^6$ source budget samples.

\subsection{Unpaired one-to-many physics translation maps}
With this set of experiments, we want to assess the capabilities of the method to generalize and learn stochastic map solutions. We increased the number of target sample size and source budget simulations for these more complex scenarios, more detailed information is provided in the Appendix \ref{app:exps}.

\textbf{Stochastic Pendulum}. We extend the same ODE of \Eqref{eq:pendlum-fric}, but in a stochastic setting. Each $\vect x$ simulation of the imperfect model $f_p$, presents multiple $\vect y$-s mapping solutions. The samples from the DGP model come with the $\xi$ fiction term as a stochastic parameter. To analyse the capabilities of the learned model, we used two configuration settings:
\begin{itemize}
    \item A one-to-two maps: $\xi \sim (1-\alpha) \xi_1 + \alpha \xi_2 $ for some $\alpha \in (0, 1), \xi_1,\xi_2 \in [0.6, 1.5]$
    \item A one-to-many maps: $\xi \sim \mathcal{U}(0.6, 1.5)$
\end{itemize}

\textbf{Stochastic Advection Diffusion}. This setup extends the previous advection-diffusion formulation \ref{eq:adv-diff} to a more complex scenario, where $\beta$ is sampled uniformly, $\beta \sim \mathcal{U}(10^{-2}, 0.1)$.

\subsection{Results}
Primarily, we compare our OT solution against the WGAN, which is the most popular method for unpaired translation problems. As noted by \cite{korotin2023neural}, WGAN addresses only the optimal transport cost, whereas the OT methods tackle the more challenging problem of solving optimal coupling. Furthermore, we systematically extended our experiments to compare various approaches for incorporating physics into machine learning models—ranging from grey-box methods that explicitly utilize the $f_p$ physical model to purely data-driven black-box models. To ensure fair comparison, for each task and method we implemented a comparable black-box architectures (e.g LSTM, ResNet, U-Net), matching the complexity and capabilities of their physics-informed counterparts. More details of each model architecture are reported in the Appendix \ref{app:exps}. We present our experimental results using the joint evaluation method in Table \ref{tab:grey-vs-black-box}.

\begin{figure}[h!]
    \centering
    \includegraphics[width=\linewidth, height=0.125\textheight]{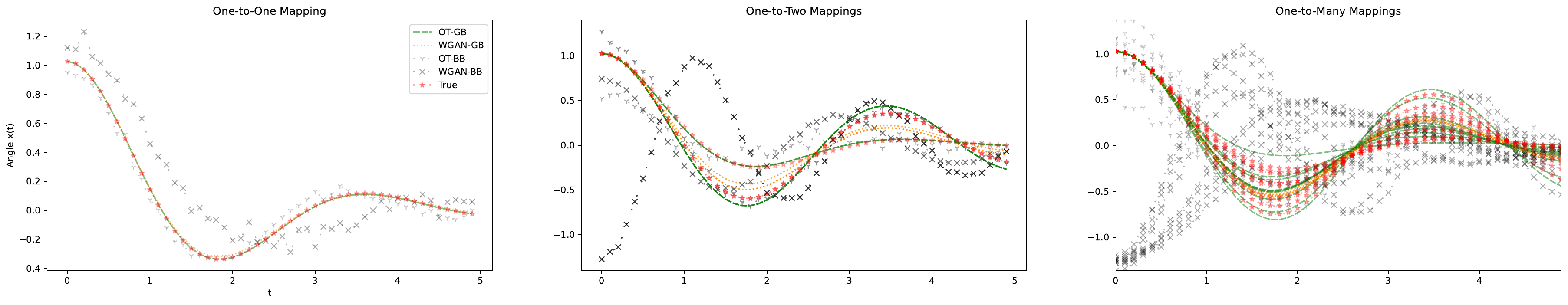}
    \caption{\small \textbf{Comparison of grey-box and black-box models on different Pendulum settings}. Black-box models fail to learn accurate conditional mappings despite $\approx100$M training source samples, while grey-box models excel with the limited simulation budget used in our experiment setting.}
    \label{fig:gb-vs-bb}
\end{figure}

By first focusing on the comparison between OT and WGAN approaches with only grey-box models (blue table headers), we find that both methods perform similarly for one-to-one translation maps. However, the OT method significantly outperforms WGAN in the stochastic setting due to the weak-OT formulation, which induces more variance while closely approximating the true distributions. Figure \ref{fig:gb-vs-bb} demonstrates this behaviour in the stochastic setting (one-to-two and one-to-many maps), where the OT grey-box model (green line) generates multi-modal maps closely aligned with the true solutions, in contrast to the WGAN method (orange dots).

Additionally, Table \ref{tab:grey-vs-black-box} demonstrates the superior performance of grey-box models compared to their black-box counterparts, with the latter showing significantly lower performance on the joint evaluation metrics. Further analysis in Figure \ref{fig:gb-vs-bb} inspects their generations for the Pendulum experiments (grey marker lines). Although black-box models capture the global problem structure, they can generate physically implausible trajectories with abrupt deviations. Furthermore, the final plot notably demonstrates anomalous trajectory generations in the initial phase, likely stemming from incorrect handling of input and conditional parameters. This suggests their difficulties in learning proper physics parameter integration within the black-box framework.


\section{Discussion}

Our experimental results demonstrate the importance of physics-informed architecture using grey-box models, showing particular strength at the generation task, and providing more reliable generations for the given input and conditional parameters. In contrast, fully data-driven approaches exhibit greater difficulty in learning the precise dynamics of physics problems and may produce non-physics-compliant generations. Furthermore, the results of the OT grey-box method highlight the effectiveness in learning conditional maps, which yields more accurate and appropriately sparse mapping solutions under stochastic settings.

\begin{figure}[ht!]
  \centering
  \includegraphics[width=0.8\columnwidth]{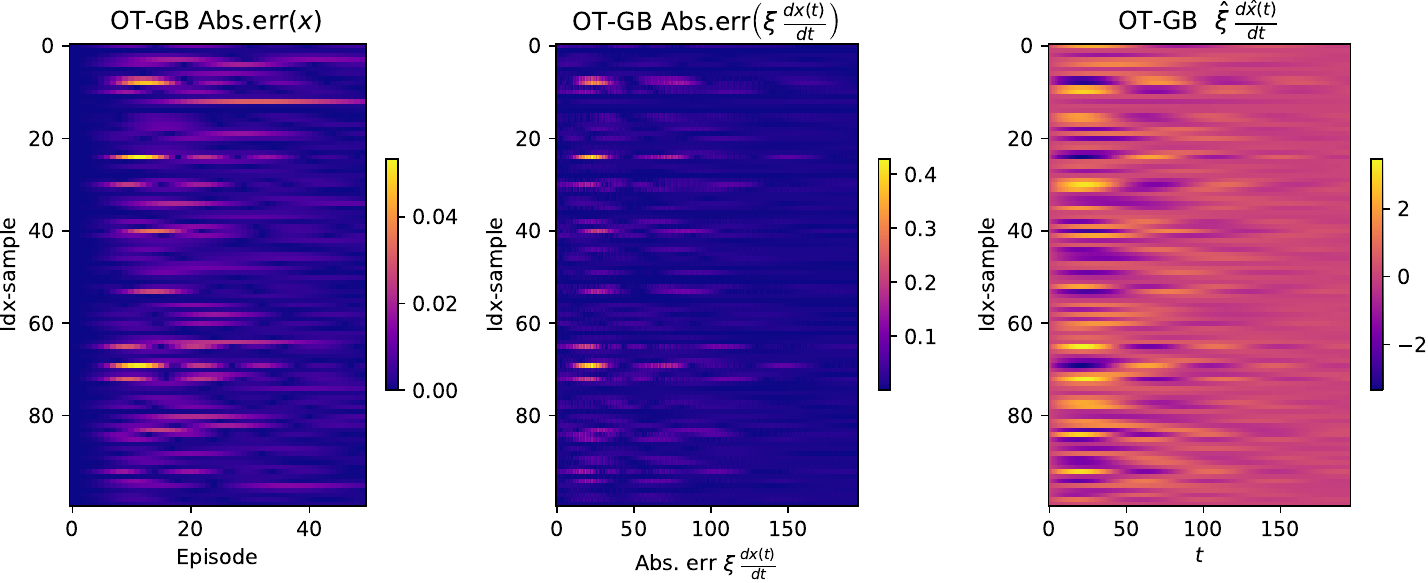}
  \caption{\small \textbf{Visualization of trajectory-level and function-level errors of the OT-GB model predictions.} The figure compares errors at different scales on the one-to-one Pendulum experiment: trajectory-level deviations (left) and in-solver function evaluation errors (center). The grey-box model exhibits low errors across both levels. The last plot (right) shows the in-solver prediction of the learned component level, which shows smooth changes and stability of the dynamics, avoiding abrupt deviations.}
  \label{fig:ot_gb_dynamics}
\end{figure}

\textbf{Inspecting inside the Grey-Box: Component Analysis}. Moreover, our approach offers some advantages in terms of interpretability, it enables granular inspection of individual components of the learned physics model. By quantifying the contributions of individual components ($f_p$ and learned components),  we gain insights into how each part influences overall model performance. This component-wise analysis surpass black-box models, allowing to examine function evaluations during continuous time evolution and optionally its gradient dynamics. In Figure \ref{fig:ot_gb_dynamics} we present a component analysis for the Pendulum one-to-one translation mapping. This analysis extends to inspect the impact of the optimal transport (OT) solution, which at the functional level comes with minimal changes of the learned components and smooth changes in the whole dynamic. These evaluations, combined with higher predictive performance, demonstrate that grey-box models enable deeper analysis and more interpretable predictions, leading to more trustworthy solutions.

\textbf{Conclusions}. These findings collectively demonstrate the advantages of our OT grey-box approach, particularly in scenarios where maintaining the physical consistency of model parameters is crucial. Moreover, our hybrid methodology maintains interpretability while achieving superior performance, positioning it as a robust framework for physics-informed machine learning applications. In fact, the grey-box component analysis studies our approach further by providing diagnostic insights into model behaviour and generation mechanisms. While they have deeper model transparency and better generation, grey-box models require differentiable physics models and may introduce computational complexity.



\section{Acknowledgments and Disclosure of Funding}
We acknowledge the financial support of the Swiss National Science Foundation SNF within the
MIGRATE project (grant no. 209434). A Multidisciplinary and Integrated Approach for geothermal exploration, CRSII5\_209434. We also acknowledge the support of the SNF project Interpretable Condition Monitoring for Complex Engineering Systems, JSPS 2025, IZLJZ2\_214000, which allowed for scientific exchanges and rich discussions with the team of Prof. Naoya Takeishi. The computations were performed at the University of Geneva on the "Baobab" and "Yggdrasil" HPC clusters.

\newpage
\bibliography{iclr2025_conference}
\bibliographystyle{iclr2025_conference}
\newpage
\appendix

\section{Background}
\label{app:background}
\subsection{Optimal Transport}
\label{app:background-ot}
\textbf{The Monge problem}. Given two measures $\mu, \nu \in \mathcal{P}(\mathbb{R}^d)$, and a ground cost function $c: \mathcal{X} \times \mathcal{Y} \mapsto \mathbb{R}$, Monge's formulation defines the deterministic map $T$ that pushes the source probability measure $\mathcal{X}$ to the target probability measure $\mathcal{Y}$ which minimizes the total cost of transport. Formally:
\begin{equation}
    T^* = \arginf_{T_{\#}\mu=\nu} \int_{\mathcal{X}} c(\vect{x}, T(\vect x)) d\mu(\vect x),
\end{equation}
where $T_{\#}$ defines the pushforward operator. 
However, this formulation of the optimal transportation problem does not fit all settings because it does not allow mass splitting \cite{book_villani_2008}.  For example, it cannot handle the case where $\mu$ is a Dirac distribution and $\nu$ is a non-Dirac distribution. 

\textbf{Kantorovich relaxation}. A major advancement in addressing the problem was made by Kantorovich's formulation, where instead of the deterministic map established by Monge it proposes a probabilistic approach allowing the transportation of mass from a single source point to various target points (mass splitting), resulting in: 
\begin{equation}
\label{eq:strong-ot}
W(\mu, \nu) = \inf_{\pi \in \Pi(\mu,\nu)} \int_{\mathcal{X} \times \mathcal{Y}} c(\vect{x}, \vect{y}) d\pi(\vect{x}, \vect{y)},
\end{equation}
where $\Pi(\mu, \nu)$ is the collection of all probability measures on $\mathcal{X} \times \mathcal{Y}$ with marginals $\mu$ on $\mathcal{X}$ and $\nu$ on $\mathcal{Y}$, and the  $\pi^*$ of the $W(\mu, \nu)$ is called the optimal transport plan. \Eqref{eq:strong-ot} admits a dual formulation, as shown in \cite{book_villani_2008} Sect. 5, which comes in the form of the following constrained concave maximization problem:
\begin{equation}
	W(\mu, \nu) = \sup_{f \in L^1(\nu)} \int_\mathcal{Y} f(\vect{y}) d\nu(\vect{y}) + \int_\mathcal{X} f^c(\vect{x})  d\mu(\vect{x}),
\end{equation}
where  $f: \mathcal{Y}\rightarrow \mathbb{R}$ and  $f^c(\vect{x}) = \text{inf}_{\vect{y}} \left[c(\vect{x}, \vect{y}) - f( \vect{y})\right]$ is called the $c$-transform of $f$. 

\textbf{Weak OT formulation}. A generalization of the optimal transport formulations comes with the Weak OT formulation \cite{gozlan2017}:
\begin{equation}
\label{eq:weak-ot}
    W(\mu, \nu) = \inf_{\pi \in \Pi(\mu,\nu)} \int_{\mathcal{X}} C(\vect{x}, \pi(\cdot | \vect{x}))d\pi(\vect{x}),
\end{equation}
where $C: \mathcal{X} \times \mathcal{P}(\mathcal{Y}) \rightarrow \mathbb{R}$ is a cost function that takes a point and a distribution as input. For cost $C(x, \nu) = \int_\mathcal{Y} c(\vect{x},\vect{y}) d\nu(\vect{y})$, the Weak OT in \Eqref{eq:weak-ot} recovers the classic formulation \ref{eq:strong-ot}. If the costs $C(\vect{x}, \nu)$ is lower bounded, convex in $\nu$, jointly lower semi-continuous in an appropriate sense, \cite{backhoff-2019} proved that the minimizer $\pi^*$ always exists. An example of weak OT cost, used in our experiments, for $\mathcal{X}=\mathcal{Y}\subseteq \mathbb{R}^d$ is the $\gamma$-weak ($\gamma \ge 0)$ Wasserstein-2 ($W_{2,\gamma}$): 
\begin{align}
\label{eq:weak-ot-2}
\begin{split}
    W_{2,\gamma}(\mu, \nu) &= \inf_{\pi \in \Pi(\mu,\nu)} \int_{\mathcal{X}} C_{2,\gamma}(\vect{x}, \nu)) d\pi(\vect{x}) \\
    C_{2,\gamma}(\vect{x}, \nu) &= \int_\mathcal{Y} \frac{1}{2}\left\vert\left\vert \vect{x}-\vect{y} \right\vert\right\vert^2 d\nu(\vect{y}) - \frac{\gamma}{2} \Var (\nu) \; \\
    \Var (\nu) &= \int_\mathcal{Y} \left\vert\left\vert \vect{y}- \mathbb{E}_{\vect{y'}\sim\nu}\left[\vect{y'}\right] \right\vert\right\vert^2 d\nu(\vect{y}),
\end{split}
\end{align}
where $\Var(\nu)$ is concave and non-negative, making the cost appropriate for any $\gamma \in [0,1]$, thus the minimizer $\pi^*$ always exists. Moreover, for any appropriate cost of \Eqref{eq:weak-ot}, it admits the following dual problem:
\begin{align}
\label{eq:weak-ot-dual}
\begin{split}
    W(\mu, \nu) &= \sup_{f} \int_\mathcal{Y} f(\vect{y}) d\nu(\vect{y}) + \int_\mathcal{X} f^C(\vect{x})  d\mu(\vect{x}) \\
     f^C &= \inf_{\vect{\nu} \in \mathcal{P}(\mathcal{Y})} \left\{ C(\vect x, \nu) - \int_{\mathcal{Y}} f(\vect{y}) d\nu(\vect{y}) \right\},
\end{split}
\end{align}
with $f$ being upper-bounded continuous function with not very rapid growth $f^C$ is the weak $C$-transform of $f$ \cite{backhoff-2019}.

Through its flexible framework, weak-OT generalizes various optimal transport problems via appropriate weak cost selection, including the  Schr\"{o}dinger/entropic transport problem, martingale transport, and semi-martingale transport problems \cite{backhoff-2020}. This generalization not only enables more efficient optimization but also induces stochastic solutions to the classical OT formulation, an essential characteristic for addressing one-to-many mapping solutions as needed in our setting.


\subsection{Deep Grey-Box Generative Models}
\label{app:background-gb}
Deep grey-box models combine two components: theory-based models (e.g a physics model) $f_p$ and data-driven deep neural networks $f_\psi$. They can be expressed as a functional $ \mathcal{T}(f_p, f_\psi; \vect x)$ that takes as input the two models as well as their input, and outputs $\vect y \in \mathcal{Y}$. In this paper, we consider $f_p$ as a physics model that takes in input $\vect x \in \mathcal{X}$ and some physics parameter $\vect \theta \in \Theta$ of the system, $f_\psi$ additionally takes some latent variable $\vect z \in \mathcal{Z}$, to solve generative tasks and optionally also the output of $f_p$. $\mathcal{T}$ is defined as:
\begin{equation}
    \mathcal{T}(f_p, f_\phi; \vect x,\vect \theta, \vect z) = \text{ODESolve}\left[\frac{d\vect y(t)}{dt} = f_\phi(\vect{y}(t), \vect \theta, \vect z) \circ f_p(\vect y(t), \vect \theta)\;\bigg\vert\;  \vect y(0) = \vect x(0)\right ],
\end{equation}
where $\vect x$ is the initial condition of the dynamics. The function composition $f_\phi \circ f_p$ allows the learned component to potentially override the physics model's behaviour, contrary to $f_p \circ f_\phi$ where $f_p$ cannot be ignored. The $f_\phi \circ f_p$ compositional effect has been extensively investigated by \cite{naoya_2023}, highlighting the need for proper regularisation of the learned component.

\textbf{Physics-based Neural ODEs}. Our deep grey-box models build on Neural ODEs \cite{chen2018neuralode}, which use neural networks to learn time derivatives through continuous transformation of input data. This approach is enabled by efficient backpropagation through ODE solvers and providing tractable adjoint methods for parameter updates. We adopt this framework for hybrid modelling, solving:
\begin{align}
\label{eq:phys-neural-ode}
\mathcal{T}(f_p, f_\phi; \vect x,\vect \theta, \vect z) = \vect{y}(0) + \int_{t_0}^{t_1}  f_\phi(\vect{y}(t), \vect \theta, \vect z) \circ f_p(\vect y(t), \vect \theta) dt \;.
\end{align}




\section{Experiments configuration}
\label{app:exps}
In the following subsections, we describe each setting used in our experiments. Implementation details and reproducible experiments are available in our public GitHub repository. For each task, the learned conditional mapping network $T_\phi$ includes a learnable encoder of the conditional parameter $\vect \theta$ whose output is concatenated to the input of each hidden layer. In addition, for the stochastic configuration setting, the mapping network takes as input a latent noise $\vect z$, with the same dimension of the network input, sampled from a Categorical or Uniform distribution depending on the task solved.

\subsection{Damped Pendulum}
The initial pendulum's angle and angular velocity sampled from a uniform distribution, respectively  $\omega \sim \mathcal{U}(0.785, 3.14)$, $x(0) \sim \mathcal{U}(-1.57, 1.57)$. Generation of source and target samples are carried out using Runge-Kutta method of order 4 (RK-4), with time interval $[0,50]$ and $\Delta t=0.1$, producing  trajectory of dimension $\vect x, \vect y \in \mathbb{R}^{50}$. The target samples are generated from the complete physics model:
\begin{equation}
    \underbrace{\frac{\mathrm{d}^2 }{\mathrm{d} t^2} x(t)+\omega^2 \sin x(t)}_{f_p} + \xi\frac{x(t)}{dt}=0,
\end{equation}
The source  distribution samples are generated by the friction-less pendulum model $f_p$.

\textbf{One-to-one translation}. We generated $|Y| = 1000$ target samples with $\xi=1.2$, and used $|X| = 10^6$ samples of the source distribution as a constraints simulation budget.

\textbf{Stochastic Damped Pendulum}. The target samples from the DGP model come with the $\xi$ fiction term as a stochastic parameter. To analyse the capabilities of the methods, we used two configuration settings:
\begin{itemize}
    \item A one-to-two maps: $\xi \sim (1-\alpha) \xi_1 + \alpha \xi_2 $ for some $\alpha \in (0, 1), \xi_1,\xi_2 \in [0.6, 1.5]$, with $|Y|=1800$ target samples, $|X| = 2 \cdot 10^6$ simulations budget.
    \item A one-to-many maps: $\xi \sim \mathcal{U}(0.6, 1.5)$, with $|Y|=3000$ target samples, $|X| = 4 \cdot 10^6$ simulations budget.
\end{itemize}

\textbf{Model Architectures}. The critic $f_\psi$ comes as a MLP with $[250,100,100,50]$ hidden layers taking as input the whole trajectory. The grey-box model is implemented as a NeuralODE with the same discretize solver used for generating the dataset, and the network involves a two-layer conditional MLP with 64 neurons. Instead, the black box model used in this task is implemented as an LSTM with 4 hidden layers that inputs the source trajectory $\vect x$ and its conditional $\vect \theta$ and outputs the completed trajectory $\vect y$.

\subsection{Reaction diffusion} The task involves a two-dimensional reaction-diffusion PDE of FitzHugh-Nagumo mode:
\begin{equation}
\frac{du}{dt} = a\Delta u + u - u^3 -v - k, \quad \frac{dv}{dt} = b\Delta v + u -v .
\end{equation}
$\Delta$ describes the Laplace operator, $a, b$ are the parameter of the system, and $u(0), v(0)$ the initial conditions.
To generate the source samples, the model is misspecified by the $k$ parameter, which is not included in the imperfect model. The simulations result in a $[u(t),v(t)] \in \mathbb{R}^{2 \times 32\times32}$ states' space, evaluated on $15$ episodes of the dynamic system solved by RK-4, thus $\vect x, \vect y \in \mathbb{R}^{2\times32\times32\times 15}$.
We used $|Y|=500$ target samples and $|X|=1.5 \cdot 10^6$  source budget samples.

\textbf{Model Architectures}. The neural network used by grey-box model is a NeuralODE and uses a 2D-ConvNet with the same configuration of \cite{wehenkel2023robust}. Instead, the black-box models and the critics are implemented using 3D-UNet model taking in input the whole trajectory (and its conditional in the case of the generator).

\subsection{Advection Diffusion}
To produce the dataset, we used the following PDE: 
\begin{equation}
    \underbrace{\frac{d T(s,t)}{dt} - \alpha \frac{d^2 T(s,t)}{ds^2}}_{f_p} + \beta \frac{dT(x,s)}{ds} = 0,
\end{equation}
with $s$ being a 1-D spatial dimension diffusion on 20-point even grid. We evolved the system was evolved for $50$ time steps, with a time-step size of $\Delta t=0.02$, solved by RK-4. The simulation corresponds to a temperature matrix of $\vect T \in \mathbb{R}^{20\times 50}$. The boundary conditions are as follows: $T (0, s) = c \sin(\pi s/2), \;T (t, 0) = T (t, 2) = 0$, where $c$ is randomly sampled between $[0.5, 1.5]$.
The known $f_p$ physics model of the source simulations does not account the advection coefficient $\beta$, and the $\alpha$ diffusion coefficient is sampled $\mathcal{U}(10^{-2}, 10^{-1})$.

\textbf{One-to-one translation} For the one-to-one translation map, the DGP samples include the $\beta$ term, fixed to $\beta=0.1$, with $|Y|=1000$ target samples and $|X|=10^6$ source budget samples.

\textbf{Stochastic Advection Diffusion}. We increased the number of target sample size and source budget simulations to $|Y|=7500$ and $|X|=4 \cdot 10^6$. This setup extends the previous advection-diffusion formulation \ref{eq:adv-diff} to a more complex scenario, where $\beta$ is sampled uniformly, $\beta \sim \mathcal{U}(10^{-2}, 0.1)$.

\textbf{Model Architectures}. The neural networks used by grey-box model is a NeuralODE and uses  a 1D-ConvNet with 2 layers, 4 filters and a kernel size of 3, at training time we used an Euler solver for fast compute. Instead, the black box model and the $f_p$ critic are implemented as a ResNet with 1D convolution.

\clearpage
\section{Method}
\label{app:alg}
\begin{algorithm}[h]
    \SetAlgorithmName{Algorithm}{empty}{Empty}
    \SetKwInOut{Input}{Input}
    \SetKwInOut{Output}{Output}
    \Input{Incomplete physics $f_p$; distributions: $p(\vect \theta), \nu(\vect y)$; Potential network: $f_\psi$; mapping network: $T_\phi$; Deep Grey-Box model:$\mathcal{T}_\phi(\cdot \;; f_p, T_\phi)$; weak-cost $\hat{C}_{k,\gamma=1}$, gradient penalty function: $g_p$, penalty term: $\lambda=1$}
    \Output{Grey-box model $\mathcal{T}_\phi(\cdot \;; f_p, T_\phi)$}
    \Repeat{not converged or exceeded simulation budget}{
    \For{$i = 1, 2, \dots, T_{\text{max}}$}{
        \textbf{sample batch: $\{ (\vect x_0, \vect \theta) \} \sim p(\vect x_0, \vect \theta)$} \\
        \textbf{simulate: $\{(\vect x, \vect \theta)\} \leftarrow \{ S_{f_p}(\vect x_0, \vect \theta) : \{ (\vect x_0, \vect \theta) \} \}$}\\
        \textbf{sample batch: $\{(\vect x, \vect \theta, \vect z)\} \leftarrow \{ \vect z \sim \eta : \{(\vect x, \vect \theta)\} \}$} \\
        \textbf{generate: $\{\hat{\vect y}, \vect x, \vect \theta, \vect z \} \leftarrow \left\{ \mathcal{T}(\vect x, \vect \theta, \vect z; f_p, T_\phi):\{(\vect x, \vect \theta, \vect z)\}\right\}$}\\
        
        $\mathcal{L}_{T_\phi} \leftarrow \frac{1}{|\{\vect x \}|} \sum_{\vect x} \left[ \hat{C}_{k,1}(\vect x, \hat{\vect y}) - \frac{1}{|\{\vect z\}|} \sum_{z} f_\psi (\hat{\vect y}) \right]$ \\
        \textbf{update $\phi$:} $\frac{\partial \mathcal{L}_T}{\partial \phi}$ \\
    }
    
    \textbf{sample batch:} $\{ \vect y \} \sim \nu(\vect y)$ \\
    $\mathcal{L}_{f_{\psi}} \leftarrow \frac{1}{|\{\hat{\vect y} \}|} \sum_{\hat{\vect y}} f_\psi (\hat{\vect y}) - \frac{1}{|\{\vect y\}|} \sum_{\vect y} f_\psi (\vect y) + \lambda (g_p(\hat{\vect{y}},\vect y)) $  \\
    \textbf{update $\psi$:} $\frac{\partial \mathcal{L}_g}{\partial \psi}$ \\
    \Return{$\mathcal{T}_\phi(\cdot \;; f_p, T_\phi)$}
}
\end{algorithm}

The solution of the Maximin problem of \Eqref{eq:kernel-maximin} is achieved using an adversarial method (similarly to WGAN \cite{wgans-2017b}), by alternating the optimization of the network parameters of the mapping network $T=T_\phi$ and the potential/critic $f = f_\psi$. To satisfy the smoothness of not very rapid growth needed by the OT formulation, we enforce the Lipschitz constraint on the potential function $f_\psi$ optimizing the gradient penalty loss $g_p$ of \cite{wgans-2017b}:
\begin{align}
    g_p(\hat{\vect{y}}, \vect y) = \mathbb{E}_{\tilde{\vect{y}} \sim \tilde \nu}\left[ \left( \left\| \nabla_{\tilde{\vect y}}f_\psi ( \tilde{\vect{y}}) \right \|_{2} - 1 \right)^2 \right],
\end{align}
where $\tilde{\vect{y}} \sim \tilde{\nu}(\tilde{\vect{y}})$ is sampled uniformly along straight lines between pairs of points sampled from the target distribution $\vect y \sim \nu$ and the generator distribution $\hat{\vect{y}}\sim \hat{\nu}$.

Practically, when solving the weak OT cost we require a stochastic estimation of cost function $C_{k, \gamma}$ of  \Eqref{eq:kernel-cost}, \cite{korotin2023kernel} proposed the following estimation using the unbiased Monte-Carlo estimator with $|Z|\ge2$:
\begin{align}
\begin{split}
    \hat{C}_{k, \gamma}(\vect x, \mathcal{T}_{\vect x}(Z)) &= \frac{1}{2} k(\vect x, \vect x) + \frac{1-\gamma}{2 |Z|} \sum_{\vect z \in Z} k(\mathcal{T}_{\vect x}(\vect z), \mathcal{T}_{\vect x}(\vect z)) - \\ -&   \frac{1}{|Z|}\sum_{\vect z \in Z}  k(\vect x, \mathcal{T}_{\vect x}(\vect z)) + \frac{\gamma}{2|Z|(|Z|-1)}\sum_{\vect z \neq \vect{z'}}  k(\mathcal{T}_{\vect x}(\vect z), \mathcal{T}_{\vect x}(\vect z')),
\end{split}
\end{align}
which comes with a quadratic time complexity $\mathcal{O}(|Z|^2)$ for computing the variance estimate of all pairs $\vect z, \vect z'$ in the batch. In our experiments, the kernel function used is the distance-based kernel that computes the Frobenius norm distance between pair samples.

\end{document}